\documentclass[twoside]{article}
\usepackage[preprint,nohyperref]{jmlr2e}
\usepackage[utf8]{inputenc}
\usepackage[T1]{fontenc}
\usepackage{textcomp,amsmath,amssymb}
\usepackage{graphicx,booktabs,array,xltabular,adjustbox,float,wrapfig}
\usepackage{flafter,placeins,needspace}
\usepackage{xcolor,microtype,enumitem}
\usepackage[font=small,labelfont=normalfont,labelsep=colon,format=hang,skip=7pt]{caption}
\usepackage[hyphens]{url}
\usepackage[colorlinks=true,linkcolor=linkblue,citecolor=linkblue,urlcolor=linkblue]{hyperref}
\usepackage[capitalize,noabbrev,nameinlink]{cleveref}
\usepackage[most]{tcolorbox}
\definecolor{linkblue}{rgb}{0,0,0.5}
\definecolor{opal}{HTML}{1F57B5}
\definecolor{muted}{gray}{0.4}
\newcolumntype{L}[1]{>{\raggedright\arraybackslash}p{#1}}
\newcolumntype{C}[1]{>{\centering\arraybackslash}p{#1}}

\newcommand{\foldhead}[1]{\par\medskip\noindent\textbf{#1}\par\nopagebreak\smallskip}
\newcommand{\exhead}[1]{\par\Needspace{6\baselineskip}\noindent\textbf{#1}\par\nopagebreak\smallskip}
\newcommand{\agentquote}[2]{\par\smallskip{\small\color{muted}#1}\par\nopagebreak{\small\itshape #2}\par\smallskip}
\definecolor{theoFrame}{HTML}{2B6CB0}
\newtcolorbox{examplebox}[1]{enhanced,rounded corners,arc=3pt,
  boxrule=1.1pt,coltitle=black,fonttitle=\bfseries,
  before skip=10pt,after skip=10pt,
  attach boxed title to top left={xshift=0.8em,yshift*=-\tcboxedtitleheight/2},
  boxed title style={sharp corners,boxrule=0pt,colback=theoFrame!25},
  colback=theoFrame!7,colframe=theoFrame!85,title={#1},
  left=9pt,right=9pt,top=9pt,bottom=7pt}
\setlist{itemsep=2pt,topsep=4pt}

\hypersetup{pdftitle={Reactive Optimizers},pdfauthor={Xingjian Bai, Jia-Hao Liang, Jayden Teoh, Vincent Sitzmann}}
\title{\textbf{Reactive Optimizers}}
\author{\centering Xingjian Bai\quad Jia-Hao Liang\quad Jayden Teoh\quad Vincent Sitzmann\\[4pt]\makebox[\linewidth][c]{\normalfont\small Massachusetts Institute of Technology}}
\date{}
\ShortHeadings{}{}
\firstpageno{1}
\begin{document}
\maketitle
\begin{abstract}
We study a reactive optimizer that uses a pretrained language-model agent to inspect training dynamics and rewrite executable optimizer code during a single training run. The agent can change update rules and schedules for individual parameter groups, without optimizer-specific meta-training or memory transferred from other runs. We evaluate this approach, OPAL, on language modeling and image generation when a recipe has not been tuned for the target setting. Across the evaluated settings, it matches or improves on selected fixed recipes. Analysis of its decisions shows changes to learning-rate schedules, optimizer structure, numerical precision, and momentum. Controlled experiments also show adaptation to repeated data corruption and reduced overfitting. We report these behaviors together with the agent's computational cost and implementation requirements.
\end{abstract}
\begin{center}\small
This document is a static version of our blog post. We recommend the original interactive version at \url{https://xingjianbai.com/reactive-optimizers/}.\\
We release the code at \url{https://github.com/xingjian-bai/opal}.
\end{center}
\medskip
\section*{Introduction}

Today, the recipe used to optimize a deep neural network is typically completely determined before training even begins. A great deal of human ingenuity has gone into priors regarding how training should progress: adaptive step sizes \citep{adam}, preconditioning \citep{shampoo,soap}, orthogonalized updates \citep{muon,moon}, schedules tied to the training step \citep{sgdr,wsd}, and learning rates scaled with model size \citep{mup}. Automated discovery \citep{ais,pi,lion,velo} searches across many runs for better optimizers. Whether hand-crafted or automatically discovered, the optimizer then runs unchanged through the end of training, adapting, if at all, according to rules written in advance. This scope leaves the following questions unanswered: How should an optimizer respond to an unexpected loss spike or stall? What if one module's gradients start to oscillate while the others keep improving smoothly? And when a new type of architecture trains poorly with Adam or Muon, is the architecture worse, or has it not yet met an optimizer suited to it?

In this work, we attempt to expand the scope of conventional optimizers and make them reactive to the training itself, leading to an ever-adapting optimizer inside a single training run. This reactive optimizer reasons on the spot and writes the weight update rule for each parameter group. Directly writing optimization code, it has the freedom to implement complicated schedules, respond to training accidents, and customize the optimizer for the single training episode in which it exists. Building on earlier attempts to adapt optimizers during training \citep{velo,daniel,pbt} and to let a model direct its own fine-tuning \citep{seal}, the reactive optimizer is entirely zero-shot, with nothing learned or retained from repeated runs, and puts the optimizer's entire algorithm up for rewriting.

We start with a practical situation, with tasks from image generation and language modeling, in which a reactive optimizer can do much better than static ones. We then turn to the emergent properties of the reactive optimizer, including sophisticated scheduling, suppressing loss spikes, and escaping overfitting. Lastly, we ask what all this costs and how practical it is at scale. We expect the reactive optimizer to grow only stronger with each generation of agents, to be most useful exactly where no recipe exists yet (for a new architecture or a new dataset), and to be one step toward solving the puzzles of deep learning optimization.

\section{Training without a recipe}\label{sec:results}

We first consider a practical setting in which the training budget permits one substantial run but not a broad search over optimization recipes. For example, suppose a PhD student has designed a new model or selected a new dataset for a transformer, with one GPU for a day and a modest budget for LLM API calls. A conventional approach is to adopt a standard recipe or one from a related task; if that recipe is unsuitable, training may exhibit loss spikes or stalled progress.

The underlying question is broader: when bounded by GPU time, researcher time, or wall-clock time, how can a deep learning model be trained to its best performance without knowing the optimal optimization recipe in advance? New architectures, modalities, languages, and scientific data all fall into this setting, where few people can afford a sweep at scale.

A reactive optimizer offers another way to use that budget: it spends LLM tokens in place of GPU/human hours. With optimizers reacting to training incidents and adjusting themselves, much of the burden of recipe tuning can be reduced.

\subsection*{Results in brief}\label{sec:brief}

We start from the optimization track (Track 3) of modded-nanogpt \citep{mng}, a language-model pre-training task (training a 162M GPT on FineWeb) with many recipes tuned specifically for this one setting. From this task, we create two slightly differing natural settings: (1) the same model trained on an alternate language dataset (we select Chinese, Thai, and Kyrgyz), and (2) the original setup with a quarter of the batch size. Neither comes with a recipe tuned for that setting, so we compare the reactive optimizer with the natural options one would have if they wanted to one-shot the new training: adopting five leaderboard recipes with natural adjustments. In all these new settings, the reactive optimizer outperforms or matches the best baseline (Figure 1).

We also apply the reactive optimizer directly to Goldfish \citep{goldfish}, which proposes a dataset of 350 languages and trains one small GPT per dataset with a standard recipe not tuned for each language. This makes it an ideal ground to test the optimizer's adaptiveness in one shot. On the Amharic dataset, the official recipe ends at 83.48 bits per sentence on held-out FLORES-200 text (lower is better); the reactive optimizer, initialized with this same recipe, ends at 81.11 in one shot (on Swahili, we obtain a similar gain). Sweeping the learning rate of the standard recipe finds a static recipe that ends ahead of the reactive optimizer, at 80.89, but the reactive optimizer initialized from that swept recipe ends ahead again, at 80.73.

We also try the reactive optimizer on image generation, a direction where the literature usually does not sweep optimization hyperparameters, since diffusion-model design and optimizer design are viewed as mostly orthogonal\footnote{For example, the standard recipes of DiT and SiT \citep{sit} train with AdamW at one constant learning rate. That has begun to change, with Muon and its variants taken up for diffusion transformers (CMuon \citep{cmuon}; \citet{muondit}; RAEv2's GMuon \citep{rae}) and a recent controlled benchmark of optimizers for diffusion models \citep{diffbench}.}. It ends ahead of the official recipe on LightningDiT-XL/1 (\citet{ldit}; FID-50K 1.83 against 2.18 with guidance, 4.33 against 5.15 without), whose official recipe trains with AdamW, and on RAEv2-B (3.19 against 3.39), whose recipe trains with a Muon variant.

Figure 1 shows these runs, each with Claude Opus 5.5 powering the agent; we name our reactive optimizer OPAL (OPtimizer with an Agent in the Loop). Together, these results show that in one-shot settings, a reactive optimizer can lead to a better solution than adhering to a predetermined recipe.

\begin{figure}[H]
\centering
\includegraphics[width=1.0\linewidth]{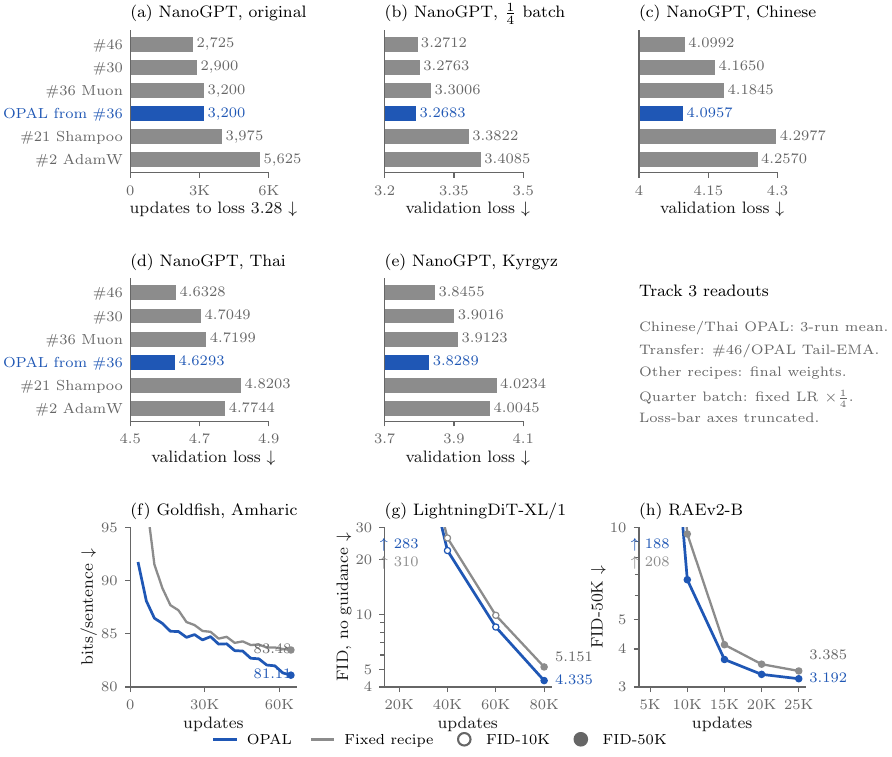}
\setcounter{figure}{0}\caption{OPAL (OPtimizer with an Agent in the Loop), our reactive optimizer, across different settings. In most settings without heavily tuned optimization hyperparameters, reactive optimizers improve the training loss and test performance upon the canonical choices. In modded-nanogpt, \#2 is AdamW, \#21 is Shampoo, \#36 is a tuned Muon, and \#30 and \#46 are more elaborate Muon-based recipes.}\label{fig:1}
\end{figure}

Beyond the practical benefit, we are intrigued by the spontaneous decisions made by the zero-shot agent, after observing and analyzing the training dynamics in its own way. In the next two sections, we explore what the reactive optimizer chooses to watch and what it changes in response.

\section{Updating with spontaneous analysis}\label{sec:loop}

\begin{wrapfigure}{r}{0.43\linewidth}
\centering
\includegraphics[width=\linewidth]{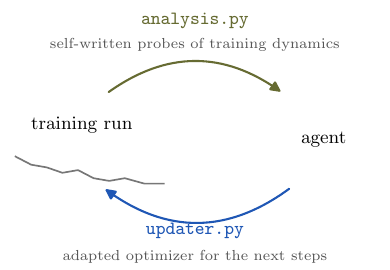}
\setcounter{figure}{1}\caption{The two programs the agent writes and rewrites iteratively during training. With code as the entire interface, we give the agent maximal freedom to choose what to watch and what to change.}\label{fig:2}
\end{wrapfigure}

Figure 2 shows how an agent updates the optimizer code on the fly. Concretely, we allow the agent to rewrite a file, \texttt{updater.py}, used in every step of optimization. We use code as the entire interface since this allows the agent to easily alter any part of the optimization. Yet, before the agent can make changes, it must observe, on its own, how training is progressing, leading to the second program it can rewrite, \texttt{analysis.py}.

What information should the agent use to make its decisions? Raw training states would be impractical: the weights, activations, and gradients would swamp its context. Instead, the statistics and geometry of training dynamics could be helpful for decisions, such as the gradient norm of each parameter group, the size of an update next to the weights it changes, the cosine between consecutive gradients, and which tensors exhibit unusually large gradient spikes. Therefore, as with the optimizer, we give the agent the most freedom: it writes a program, \texttt{analysis.py}, to extract whatever it wants to know from the training dynamics. Its self-written probes run throughout training, taking from the state of training whatever the agent asks for and handing it to the agent for its next decision.

We complete the bidirectional loop. Through \texttt{analysis.py}, training reaches the agent in whatever form the agent chooses to see it, and through \texttt{updater.py}, the agent's decisions reach every step of training. The two programs let the agent devise new optimization algorithms along the way with maximum flexibility.

\section{What is dynamically changing during optimization?}\label{sec:what}

In this section, we look at what the reactive optimizer does throughout training, examining the decisions behind these results. As a summary, Figure 3 shows when, how, and why each agent edits its code.

\begin{figure}[!t]
\centering
\includegraphics[width=1.0\linewidth]{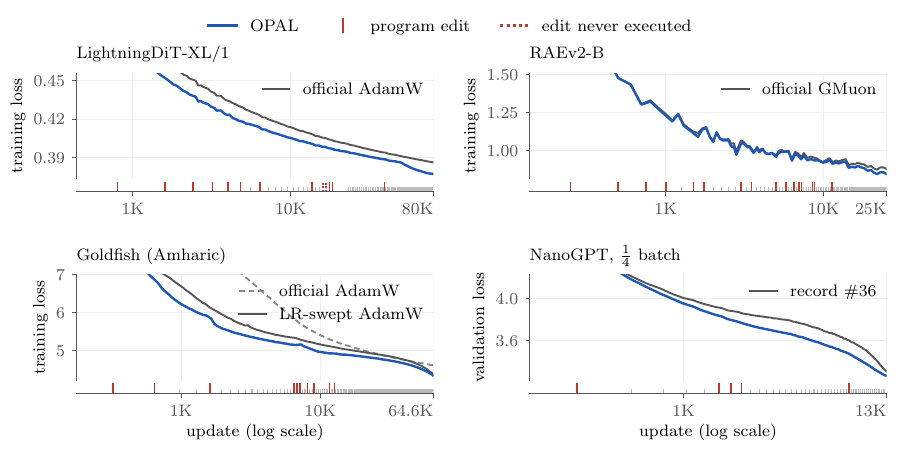}
\setcounter{figure}{2}\caption{Training trajectories and code edits in four OPAL runs. Blue curves show OPAL's loss alongside the corresponding fixed-recipe baselines; ticks below each curve mark decisions and distinguish code edits. The edit patterns differ across runs, reflecting module-specific changes made at different stages of training. The curves show the outcomes of the complete runs, rather than isolating the effect of each edit.}\label{fig:3}
\end{figure}

The rest of this section examines these adaptations type by type.

\subsection{Learning-rate schedule}\label{sec:sp-lr}

Frontier training runs come with elaborate hyperparameter schedules, such as the five stages of DeepSeek-V3's learning rate (\citet{dsv3}; Figure 4, upper left), with such schedules sometimes being written by humans during the training process \citep{opt}. Engineers can make only so many decisions, each weighing only so many signals. A reactive optimizer can take over this work and has no such limit: it decides frequently, and at each decision, it can access the entire history of the run. As Figure 4 shows, the schedules our agents built are intricate, with separate multi-stage schedules for each parameter group, and differ from run to run. Several of them raise a learning rate mid-run, a move rarely seen in modern schedulers, yet naturally helpful in many scenarios \citep{sgdr,clr}. Moreover, the four examples in Appendix~\ref{app:lr-examples} show the evidence they cited for their decisions.

We observe that the agents base these decisions on the diagnostics from \texttt{analysis.py}. Some examples are: the cosine between a module's consecutive gradients as a sign of overshooting (“Deduced from clues” in Appendix~\ref{app:lr-examples}); A/B testing between two learning rates every few hundred updates, comparing how fast the loss falls under each (“A-B-A test”); trial and error, with the learning rate raised, lowered, and raised again (“Trial and error”); and the trend of the loss, which makes agents start their cooldown earlier than planned (“Re-planned cooldown”). Each of these changes of schedule involves observation of the training dynamics and reasoning upon said observations.

\begin{figure}[!htb]
\centering
\includegraphics[width=1.0\linewidth]{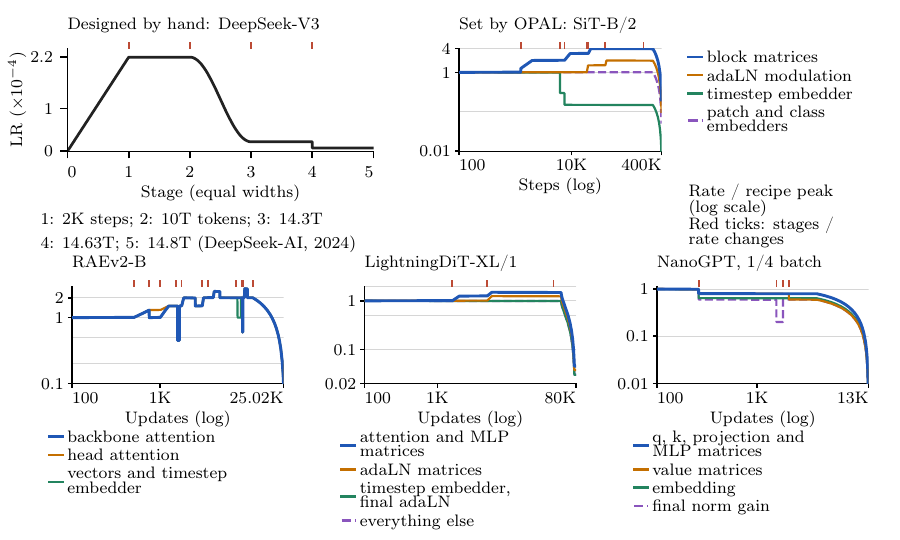}
\setcounter{figure}{3}\caption{DeepSeek-V3's learning rate, designed by hand (upper left), and learning rates OPAL set in four runs, as multiples of each recipe's rate. The schedules vary across tasks and parameter groups and include mid-run rate increases as well as cooldowns. These are observed trajectories, rather than schedules specified before training.}\label{fig:4}
\end{figure}

\subsection{Optimizer design}\label{sec:sp-opt}

The reactive optimizers also change which rule updates which tensors. In our experiments, many switch from the initially given AdamW to the well-known hybrid, Muon, for the attention and MLP matrices while remaining with AdamW for the rest. Later, they make more specific moves (Figure 5).

The optimizer can also decide where precision matters. On modded-nanogpt Track 3, it measured that the entries of the token embedding are about 15 in size, where adjacent bfloat16 values are 0.0625 apart, so as the learning rate decayed, rounding would throw away a growing share of each update. From update 1,040, it kept that specific matrix's Adam moments in float32 and wrote its weights back with compensated rounding and left every other tensor as it was.

\begin{figure}[!htb]
\centering
\includegraphics[width=1.0\linewidth]{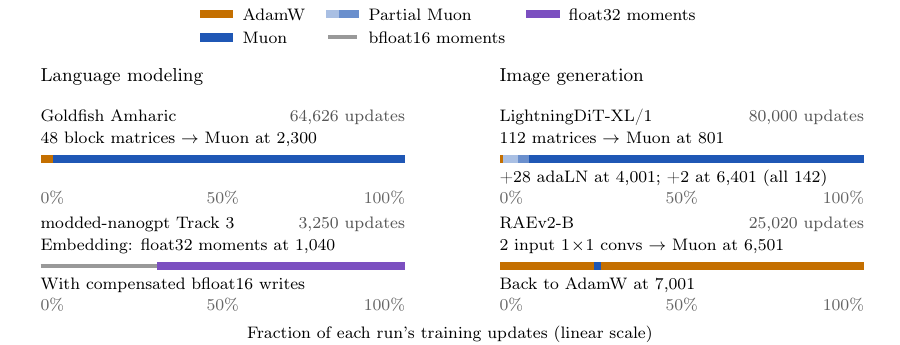}
\setcounter{figure}{4}\caption{Changes of optimizer type and of precision in four OPAL runs. In some runs, the weights are simply moved to standard Muon, while in others, OPAL moves them gradually by module type and chooses high precision for a very small part of the weights, citing the signals it observed.}\label{fig:5}
\end{figure}

\subsection{Momentum}\label{sec:sp-mom}

We investigate the reactive optimizer's ability to manipulate momentum on 2D toy landscapes, where we give the agent new abilities: resetting its momentum, returning to its best point so far, and jumping to a random nearby point. In these toy settings, standard OPAL would simply replace the default recipe with a second-order method within a few decisions and reach perfect performance very fast. We therefore restrict OPAL to the heavy-ball algorithm, while letting it control the hyperparameters. On the landscape in Figure 6, the start sits beside a nearly flat saddle where common optimizers stall. The reactive optimizer raises its momentum to cross the flat stretch, lowers it as the gradient falls near the minimum, and each time it overshoots, returns to its best point and restarts its momentum, reaching the minimum by the end of its 300-step budget. Over all 34 landscapes we ran, it achieves an average score of \textminus{}11.6 (log\textsubscript{10}(\emph{f} \textminus{} \emph{f}*), lower is better), while any single static setting does worse: \textminus{}2.2 for the best of twelve common recipes, and \textminus{}7.4 for the best of 382,080 swept recipes (see the aggregate results in the interactive blog).

\begin{figure}[!htb]
\centering
\includegraphics[width=1.0\linewidth]{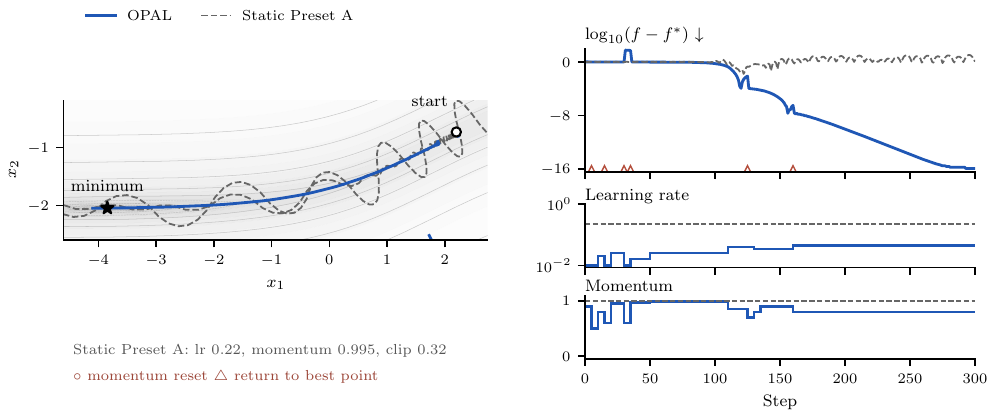}
\setcounter{figure}{5}\caption{A variant of the tilted double-well potential, one of the 34 landscapes we test optimizers on. The dashed line uses the blog's static Preset A, while the blue line traces the reactive optimizer. The reactive optimizer raises its momentum along the gentle slope, lowers it as the gradient falls, and clears it when it overshoots, reaching the optimum where common optimizers at their default settings stall.}\label{fig:6}
\end{figure}

\subsection{Loss spikes}\label{sec:sp-clip}

In our large-scale training, OPAL also limits the damage of gradient spikes with module-specific clipping: another SiT agent traced 17 of the 18 gradient spikes it logged to the timestep embedder, and added a clipping threshold for each tensor, set from that tensor's own recent gradient norm, so that a spike in the embedder no longer cuts the updates of every other tensor; its old global threshold stays only as a backstop. Neither move is standard practice, but both have precedents: clipping each tensor against its own history \citep{adagc}, and damping the one layer that spikes, as GLM-130B did for its embedding layer \citep{glm}.

Spikes can also come from the data, through the interaction of particular batches with the model state \citep{palm}, and clipping does not in general protect against noisy labels \citep{menon}. In a controlled experiment, we insert four stretches of random images and labels into CIFAR-10, and train a ResNet-18 with OPAL, which is not told about the corruption, and with static optimizers (Figure 7). The first two stretches cost OPAL 50 and 66 points of test accuracy, about as much as the baselines. In between, it added loss damping, replaced it with clipping, and then wrote a gate that holds back batches whose loss is far above recent clean ones. After it tuned that gate, the third stretch cost 12 points and the fourth none, while the clipped baselines and Stable-SPAM \citep{stablespam} still lost 28 to 76 points in each.

\begin{figure}[!htb]
\centering
\includegraphics[width=0.92\linewidth]{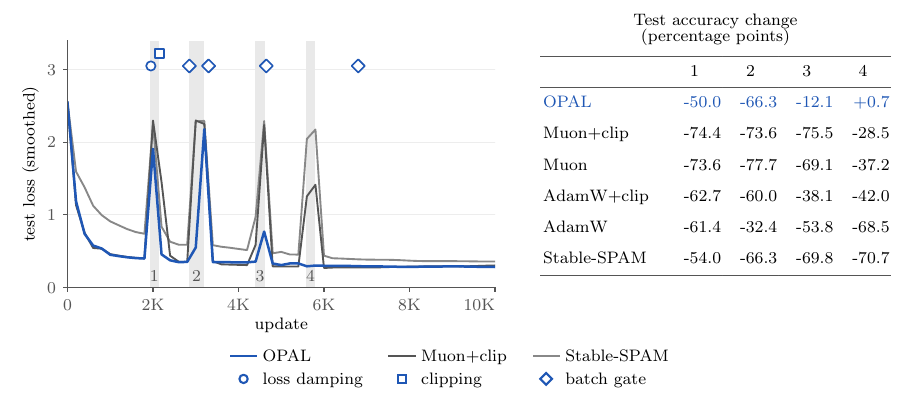}
\setcounter{figure}{6}\caption{ResNet-18 on CIFAR-10 with four stretches of random images and labels, shaded. Left: test loss; markers indicate changes to OPAL's safeguards. Right: the change in test accuracy in each stretch. Reactive agents can defend against the designed type of loss spikes, induced by data corruption, after seeing it a few times. Clean segments use Gaussian smoothing in log loss ($\sigma=300$ updates); the first 1,000 updates, corruption intervals and two evaluations after each interval remain unsmoothed.}\label{fig:7}
\end{figure}

\subsection{Overfitting}\label{sec:sp-overfit}

Training on data with wrong labels is a standard way of testing overfitting \citep{dd}. We design such a setting (Appendix~\ref{app:overfit-setting}) and ask if the optimizer can overcome overfitting by itself, reacting to signals from the training set and the validation set. Trained straight through, the AdamW baseline ends at test loss 2.49 and accuracy 73.5\% (stopped early on the validation set, at 1.01 and 72.9\%; with its weights averaged and the average chosen there, at 0.83 and 77.5\%). The reactive optimizer ends at loss 0.69 and accuracy 78.6\%. As a direct measure of memorization, when asked to classify the 4,500 mislabeled training images, the baseline gives the wrong label it was trained on for all of them, its early-stopped version for 19\% and its weight-averaged version for 14\%. The reactive optimizer, on the other hand, gives that label for only 4\%, and the true label for 78\%. Its strategy combines progressively higher learning rates with weight averaging, and it chooses its final average on the validation set, which its own analysis program calls “extended SWA” \citep{swa}.

\begin{figure}[!htb]
\centering
\includegraphics[width=0.90\linewidth]{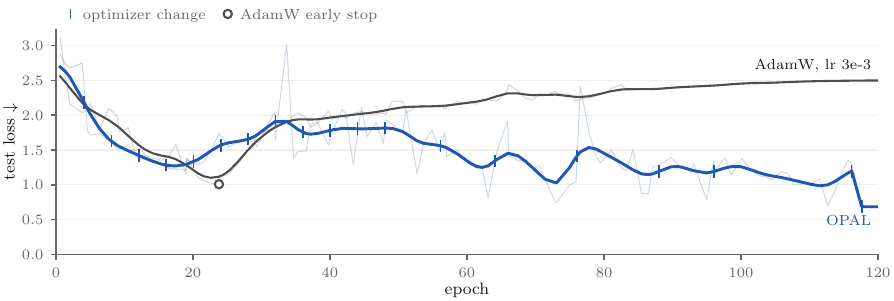}
\setcounter{figure}{7}\caption{ResNet-18 trained for 120 epochs on 45,000 CIFAR-10 images, 10\% with a wrong label. At epochs 24 and 28, validation loss climbs and the agent reacts. OPAL counters the overfitting with progressively higher learning rates and a weight average it chooses on the validation set. OPAL ends at a test loss of 0.685, compared with 2.493 for the final AdamW weights and 0.832 for its validation-selected weight average; the test set is not used by the agent.}\label{fig:8}
\par\medskip
\centering
{\small
\setlength{\tabcolsep}{5pt}
\begin{tabular}{@{}lrrrr@{}}
\toprule
 & Test loss $\downarrow$ & Test acc. $\uparrow$ & \multicolumn{2}{c}{Mislabeled training subset} \\
\cmidrule(lr){4-5}
 & & & True label $\uparrow$ & Wrong label $\downarrow$ \\
\midrule
OPAL & 0.685 & 78.6\% & 77.5\% & 3.7\% \\
AdamW, final & 2.493 & 73.5\% & 0.0\% & 100.0\% \\
AdamW, early stopping on validation & 1.010 & 72.9\% & 59.7\% & 19.4\% \\
AdamW, validation-selected EMA & 0.832 & 77.5\% & 70.0\% & 14.2\% \\
\bottomrule
\end{tabular}
}
\captionof{table}{Test performance and predictions on the mislabeled training subset. Early stopping and weight averaging use the validation set; details are in Appendix~\ref{app:overfit-setting}.}\label{tab:overfit}

\end{figure}

\subsection*{Summary of analysis}\label{sec:summary-analysis}

Above, we show that reactive optimizers can adapt their recipe to the dataset, to individual modules, and to the behavior of each run, a trait unseen in conventional optimizers. Such customization and self-adaptiveness allow them to craft sophisticated schedules and navigate complex training landscapes.

Still, tuning hyperparameters, suppressing loss spikes, and watching out for overfitting are things human experts can do, albeit less conveniently and at a higher cost. As agents get smarter, with hundreds of papers in their memory, thousands of tokens in their chain of thought, millions of training signals in their context, and every step in their hands, we are convinced that reactive optimizers will only become better, developing mechanisms during training more arcane than the ones demonstrated here, beyond what human experts could tune by hand. Such mechanisms may include tracking the second-order curvature of the loss landscape, auditing the variance of each data batch's signal, or implementing never-before-seen optimization schemes.

\section{Real-world applicability}\label{sec:cost}

Is a reactive optimizer practical? The very first concern is the time spent waiting for reasoning decisions. In its simplest design, training is stopped while the agent reasons, inducing significant overhead. In our OPAL implementation, we instead provide the option to allow training to continue while the agent thinks, and swap in the agent's edits at a later update (e.g., a median of 65 updates later on Goldfish). This should have little impact on performance, since most optimizer decisions are not especially time-critical. With this option, the runs in Figure 9 add only a little time on top of the static run (1.01\texttimes{} to 1.14\texttimes{}). At a larger scale, the current implementation still needs work to support sharded training and optimizer compilation; we leave these practical concerns to later work.

\begin{figure}[!htb]
\centering
\includegraphics[width=1.0\linewidth]{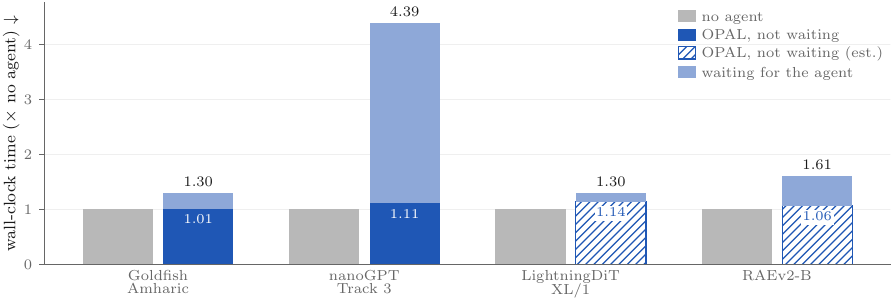}
\setcounter{figure}{8}\caption{Wall-clock time relative to the same task trained without an agent. Training that waits for the agent at every decision takes much longer (1.30\texttimes{} to 4.39\texttimes{}), while training that does not wait adds only a little time on top of the static run (1.01\texttimes{} to 1.14\texttimes{}). That remaining overhead is the cost of running what the agent wrote: an update rule that costs more per step than the recipe's (such as Muon's orthogonalization, or keeping a tensor in float32) and the measurement cost of its analysis program.}\label{fig:9}
\end{figure}

When comparing API tokens to GPU hours, the reasoning of a reactive optimizer costs little compared with the training compute. At about 100 decisions in a single training run, the LightningDiT agent costs \$6.42 at Claude Opus 5.5 list price, on top of about 74 H100-hours of training (about \$510 at AWS's current on-demand price); equipping it on larger training would only make this relative cost smaller. Admittedly, counted in FLOPs, any reactive optimizer would be impractical given the size of frontier agents. Here we simply want to point out that, for most researchers, GPU hours and token access to closed-source models are two different commodities and as the frontier continues to advance, the price of a given capability will continue to fall.

Another limitation is that the reactive optimizer needs room to observe and react; its gains shrink when a run is short and its recipe is fine-tuned. The modded-nanogpt track is such a setting: 3,250 updates of a recipe that many people have tuned for nearly this exact task; our gains there are smaller than in the other tasks we tried, which train for about 8 to 120 times as many updates. Moreover, the reactive behavior introduces another source of uncertainty, potentially making training less reproducible. Most experiments in this work are run once, a few three times; reruns on the released package are in the \href{https://github.com/xingjian-bai/opal}{OPAL repository}.

Fine-tuning platforms are bringing training to everyone. Tinker \citep{tinker} and Prime Intellect's Hosted Training \citep{pilab} let users fine-tune open-weight models for their own applications without managing the GPUs behind them. On these platforms, however, the mental burden of a good training recipe still falls on people. A reactive optimizer fits naturally on top of these platforms, offloading that burden to agents and helping each user personalize a model to their own needs.

\section{Conclusion}\label{sec:end}

Until now, optimizers have been written before training. Agents, as a cheap form of reasoner with a long context window and expert knowledge, make something new possible: an optimizer that keeps deciding while the network trains in reaction to optimization events. We have demonstrated that such a reactive optimizer watched the training dynamics and rewrote its own code as training unfolded, for different groups of parameters, at the step it deemed proper. On toy and scaled-up experiments, most of these runs ended ahead of the recipes from which they started.

The analysis and update mechanism we design is meant to be as simple and general as possible. Wrapping a standard optimizer, it assumes nothing about the model, the data, or the training dynamics. It is useful today wherever no recipe exists yet. For example, a model on a biological or other scientific dataset, an architecture for which no recipe was written, or a language for which no one has tuned a recipe are all excellent use cases for such a reactive optimizer. This is because the optimizer’s agent takes the role of a human expert at countless training steps, where no expert has ever had the time to sit, making similar decisions in response to events in the run. Seen this way, a reactive optimizer is a natural thing to build, and its gains over static ones are to be expected. Furthermore, a training run holds far more information than humans can use, almost all of which is ignored today. A reactive optimizer has all this information within reach. Imagine a world where the notion of a monolithic optimizer is gone; what replaces it is an optimization intelligence that adapts to every specific run, and it might not be gradient-based, but zeroth-order \citep{dust}, language-based or more. Perhaps, in the future, we should delegate optimization decisions to agents, not humans.

In the past decade, deep learning models have come to possess unreasonably effective power through training. Yet the very heart of their making, the optimizer, is still mechanistic. This work is an attempt to breathe the magic that training has given us into the core of training itself, forming a loop that we hope will create more capability in the future.

\section*{Acknowledgements}\label{sec:ack}

We sincerely thank \href{https://georgecazenavette.github.io/}{George Cazenavette}, \href{https://dcharatan.github.io/}{David Charatan}, \href{https://shivamduggal4.github.io/}{Shivam Duggal}, \href{https://jyopari.github.io/}{Jyo Pari} and Qianwei Jia for their invaluable advice on this write-up; thank \href{https://zhentingqi.github.io/}{Zhenting Qi} and \href{https://zqyuan.com/}{Zeqing Yuan} for their discussion. We are also grateful to Professor \href{https://web.mit.edu/phillipi/}{Phillip Isola} and the \href{https://web.mit.edu/phillipi/\#ResearchGroup}{Isola lab} for their insightful feedback on an earlier version of this work.

\FloatBarrier
\renewcommand{\refname}{References}
\setlength{\bibsep}{3pt}
\bibliography{references}

\clearpage
\appendix
\section{Related work}
Learning optimization rules and adapting them to training feedback have a long history. Early work already considered improving learning strategies themselves and learning context-dependent weight updates, through self-referential learning systems and neural fast-weight controllers \citep{schmidhuber1987selfreferential,schmidhuber1992fastweights}. \citet{andrychowicz2016learning} train recurrent networks to produce parameter updates, while \citet{li2016learning} formulate optimization as a policy learned through reinforcement learning. \citet{daniel} learn step-size controllers, and VeLO scales meta-training across optimization tasks \citep{velo}. These methods already make updates depend on the evolving training state. For reinforcement learning, Optim4RL and OPEN learn update rules tailored to its optimization difficulties \citep{lan2024learning,goldie2024can}. AdaQN instead selects a shared target online from critics trained with different hyperparameters \citep{vincent2025adaptive}. Population-based training discovers hyperparameter schedules while jointly training a population of models \citep{pbt}. SEAL uses reinforcement learning to train a language model to generate its own fine-tuning data and update directives, including optimization hyperparameters \citep{seal}. Our setting uses a pretrained general-purpose agent to inspect one ongoing run and rewrite its executable optimizer code, without optimizer-specific meta-training. The distinction lies in adapting executable update code within the run, rather than in the ability to respond to training feedback itself.

\section{Learning-rate decisions}\label{app:lr-examples}

\begin{examplebox}{Deduced from clues}

The SiT agent's analysis program measured the cosine between consecutive gradients of each parameter group. Where it was negative, the agent lowered the learning rate: the timestep embedder's went to 0.3\texttimes{} and then to 0.15\texttimes{} of the official rate, with this stated reason:

\agentquote{Step 7,506, Decision 6}{Timestep-embedder learning rate goes from 0.3\texttimes{} to 0.15\texttimes{} (1.5e-5) after update 7506, because its cosine is still around \textminus{}0.22.}

Where the cosine stayed slightly positive, for the Muon matrices, it raised the learning rate.

\end{examplebox}

\begin{examplebox}{A-B-A test}

The RAEv2 agent alternated two learning rates over consecutive stretches of a few hundred updates (A-B-A) and compared the training loss under each. After one such test, it switched to 2\texttimes{} from update 3,100, with this stated reason:

\agentquote{Update 3,000, Decision 12}{The A-B-A test points the other way from my provisional read at update 2750: 2\texttimes{} gives faster real progress than 1.5\texttimes{}.}

\end{examplebox}

\begin{examplebox}{Trial and error}

The same agent set every group's learning rate to 1.3\texttimes{}, returned the backbone's learning rate to 1\texttimes{} when the backbone's gradients began to change sign at each step, and at the next decision raised every group's learning rate to 1.5\texttimes{}:

\agentquote{Update 1,000, Decision 4}{My explanation from last time was wrong, so I'm raising the learning rate in every group to 1.5\texttimes{} (3e-4).}

It then ran learning rates of 2\texttimes{}, 2.5\texttimes{} and 2.75\texttimes{}, each for about 650 updates, and kept 2\texttimes{}. Below, the learning rate of its backbone attention matrices, as a multiple of the official rate; the two deep dips are 75-update reductions of the learning rate used as tests.

\begin{center}\includegraphics[width=0.95\linewidth]{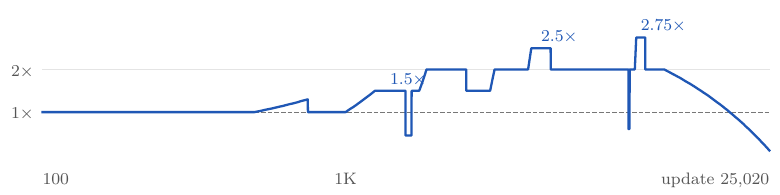}\end{center}

\end{examplebox}

\begin{examplebox}{Re-planned cooldown}

At step 190K, the SiT agent moved its planned cooldown from 320K to 280K and lowered its learning-rate floor from 0.1 to 0.05 of its peak (the bend in the SiT panel of Figure 4). Its stated reason:

\agentquote{Step 190,152, Decision 53}{the constant-rate phase is now mostly holding the weights at a noise floor set by the learning rate, rather than making progress.}

The RAEv2 agent started its cooldown at update 11,250, 45\% of the run instead of the planned 75\%, citing the trend of its loss and a paper on decaying the learning rate \citep{d2z}.

\end{examplebox}

\section{Experimental settings and decision records}

\subsection{Overfitting experiment}\label{app:overfit-setting}

A ResNet-18 trained on CIFAR-10 for 120 epochs without data augmentation. Among the 50,000 training images, 10\% are held out with their correct labels as a validation set; among the rest, 10\% get a wrong label, drawn uniformly from the wrong classes and fixed for the run. The agent saw training statistics, along with validation loss and validation accuracy; it was not told about the specific setting (that some labels were wrong), and it never saw the test performance. We make the network normalize its logits (a centered form of LogitNorm, \citep{logitnorm}) to prevent loss-hacking behavior. AdamW runs from learning rate 3e-3, which is also the starting recipe for the reactive optimizer.

OPAL's final weights are an average of its weights from epochs 96–117, chosen on the validation set. The AdamW EMA baseline uses a moving average of its weights, also chosen on the validation set.

\subsection{Settings for Figure 1}

{\small\begin{xltabular}{\textwidth}{@{}L{2.6cm}L{1.4cm}X@{}}
\toprule
Model / task & Field & Setting \\
\midrule\endhead
\bottomrule\endfoot
modded-nanogpt Track 3 & Task & the 162M GPT of the modded-nanogpt optimization track (Track 3), the same runs in each setting: official NanoGPT, the track's own (FineWeb at the official batch of 524,288 tokens per update); NanoGPT, \textonequarter{} batch (FineWeb, 13,000 updates of 131,072 tokens); NanoGPT on Chinese, Thai and Kyrgyz (the 1 GB Goldfish Chinese, Thai and Kyrgyz sets, 3,000 updates of 524,288 tokens) \\
 & Metric & official NanoGPT: the first update at which the validation loss reaches 3.28, the track's target (validation every 25 updates); \textonequarter{} batch: validation loss after 13,000 updates; Chinese, Thai and Kyrgyz: validation loss after 3,000 updates, for \#46 and OPAL from \#36 with \#46's own readout (a moving average of the weights over the last updates), for the other records with the last weights; in Chinese and Thai, OPAL's bar is the mean of three runs. The loss bars start at c, the largest multiple of 0.1 at least 0.05 below the setting's lowest value: 3.2 (\textonequarter{} batch), 4.0 (Chinese), 4.5 (Thai) and 3.7 (Kyrgyz) \\
 & Baseline & five records from the leaderboard, \#46, \#30, \#21 (Shampoo), \#36 (the tuned Muon baseline) and \#2 (AdamW), and OPAL from \#36. The leaderboard lists 2,690, 2,930, 4,100 and 3,250 updates for \#46, \#30, \#21 and \#36; \#2's bar in official NanoGPT is the leaderboard's 5,625: our own run of \#2 ended at 3.28449 after 5,625 updates and never reached 3.28; at \textonequarter{} batch every record ran with all its learning rates \texttimes{}\textonequarter{}. \\
\midrule
Goldfish, Amharic & Task & a 125M GPT trained from scratch on 1 GB of Amharic (Goldfish), 10 epochs (64,626 updates) \\
 & Metric & bits per sentence on FLORES-200 (dev + devtest) \\
 & Baseline & the official Goldfish recipe (AdamW, learning rate 1e-4, weight decay 0.01, 10\% warmup, then linear decay; gradient clipping at 1.0); nine more static recipes, each for all 10 epochs: AdamW at learning rates 6e-4, 1e-3, 2e-3 and 4e-3 with weight decay 0.1 (a four-value sweep), AdamW at 3e-4 with weight decay 0.1 or 0.01 and at 1e-4 with weight decay 0.1, and Muon on the hidden block matrices in the official recipe and at learning rate 3e-4, weight decay 0.1. OPAL ran twice: from the official recipe (at its defaults, 50 decisions; shown by default) and from the sweep's best point, AdamW at learning rate 2e-3 (200 decisions; shown with the sweep) \\
\midrule
LightningDiT-XL/1 & Task & image generation on ImageNet 256\texttimes{}256 in a VA-VAE latent space; LightningDiT-XL/1, 64 epochs \\
 & Metric & FID-50K at the end, with and without guidance; the curve shows FID-10K during training \\
 & Baseline & the official recipe, retrained by us, and the authors' released checkpoint (FID-50K 5.23 without guidance, 2.16 with) \\
\midrule
RAEv2-B & Task & image generation on ImageNet 256\texttimes{}256 in a representation-autoencoder latent space; RAEv2-B, 20 epochs \\
 & Metric & FID-50K \\
 & Baseline & the official recipe, which already trains with GMuon, a Muon variant \\
\end{xltabular}}

\Needspace{8\baselineskip}
\subsection{Program changes in Figure 3}

\foldhead{All decisions that changed a program}

\exhead{LightningDiT-XL/1}

{\fontsize{9}{10}\selectfont
\renewcommand{\arraystretch}{0.94}
\begin{xltabular}{\textwidth}{@{}>{\raggedleft\arraybackslash}p{1.1cm}L{2.2cm}X L{3.0cm}@{}}
\toprule
Step & File & What changed & Tags \\
\midrule
\endhead
\bottomrule
\endfoot
800 & \texttt{updater.py} + \texttt{analysis.py} & Moved the attention and MLP weights to Muon, 196 matrices once the fused tensors are split, each step matched to its AdamW size; kept adaLN and the rest on AdamW; planned a cosine cooldown from 60K & block-specific, time-step-specific \\
1,600 & \texttt{updater.py} + \texttt{analysis.py} & Ramped Muon's steps to 1.25\texttimes{} their matched size over updates 1,600–2,000; added a guard that cuts a step's rate when the gradient norm passes 3\texttimes{} its mean & observation-originated, time-step-specific \\
2,400 & \texttt{updater.py} + \texttt{analysis.py} & Fixed a reporting bug; split its gradient–momentum cosine by matrix type and depth, and added it for the AdamW tensors & analysis only, fix or revert \\
3,200 & \texttt{updater.py} & Added a controller that multiplies each AdamW tensor's rate by 0.9 every 100 updates while its gradient opposes its momentum (down to 0.2) & niche practice, block-specific, observation-originated \\
4,000 & \texttt{updater.py} & Removed that controller; moved the adaLN weights to Muon, each split into its six 1152\texttimes{}1152 chunks & fix or revert, niche practice, block-specific, observation-originated \\
4,800 & \texttt{updater.py} & Ramped Muon's steps to 1.5\texttimes{} their matched size for attention and MLP and 1.25\texttimes{} for adaLN, over updates 4,800–5,600 & block-specific, observation-originated, time-step-specific \\
6,400 & \texttt{updater.py} & Moved three more matrices to Muon: the timestep embedder's second layer and the final adaLN's shift and scale & block-specific, observation-originated \\
13,600 & \texttt{updater.py} & Replaced the planned cooldown with a 1\textminus{}$\surd$ decay from 54K to 0.03\texttimes{} at 76K (was cosine from 60K) & time-step-specific \\
16,000 & \texttt{analysis.py} & Added per-tensor gradient norms outside the blocks and a spike log; never applied (its answer reached the host too late) & analysis only, observation-originated \\
16,800 & \texttt{analysis.py} & Wrote the same diagnostics again; also never applied & analysis only, observation-originated \\
17,600 & \texttt{analysis.py} & Wrote them a third time; these ran & analysis only, observation-originated \\
18,400 & \texttt{updater.py} + \texttt{analysis.py} & Made the spike guard ignore the timestep embedder's first layer, whose noisy gradient had caused nearly all its cuts & fix or revert, block-specific, observation-originated \\
39,200 & \texttt{updater.py} & Moved the cooldown start from 54K to 50K, as progress at the constant rate had nearly stopped & observation-originated, time-step-specific \\
\end{xltabular}}

\exhead{RAEv2-B}

{\fontsize{9}{10}\selectfont
\renewcommand{\arraystretch}{0.94}
\begin{xltabular}{\textwidth}{@{}>{\raggedleft\arraybackslash}p{1.1cm}L{2.2cm}X L{3.0cm}@{}}
\toprule
Step & File & What changed & Tags \\
\midrule
\endhead
\bottomrule
\endfoot
250 & \texttt{updater.py} + \texttt{analysis.py} & Re-implemented the official GMuon and AdamW with a rate multiplier per parameter group (all 1\texttimes{}); planned a linear cooldown from update 18,765 to 0.05\texttimes{} & time-step-specific \\
500 & \texttt{updater.py} + \texttt{analysis.py} & Ramped every rate to 1.3\texttimes{} over updates 500–750; added a clip at twice the running gradient norm & time-step-specific \\
750 & \texttt{updater.py} + \texttt{analysis.py} & Put the backbone back to 1\texttimes{} and kept the head at 1.3\texttimes{}, after the backbone's gradients began to flip sign each step & block-specific, observation-originated \\
1,000 & \texttt{updater.py} + \texttt{analysis.py} & Raised every group to 1.5\texttimes{} over updates 1,000–1,250; its analysis now multiplies the rate by 0.8 when the loss rises 3\% above its recent best & observation-originated, time-step-specific \\
1,500 & \texttt{updater.py} + \texttt{analysis.py} & Scheduled a probe: the rate at 0.3 of its value for updates 1,576–1,650, to measure the step-size noise in the loss & niche practice, time-step-specific \\
1,750 & \texttt{updater.py} & Scheduled an A-B-A test: 2\texttimes{} for updates 1,850–2,500, between stretches at 1.5\texttimes{} & niche practice, observation-originated, time-step-specific \\
3,000 & \texttt{updater.py} & Chose 2\texttimes{} from update 3,100, after the test showed faster progress at 2\texttimes{} than at 1.5\texttimes{} & observation-originated, time-step-specific \\
3,500 & \texttt{updater.py} & Scheduled a second test: 2.5\texttimes{} for updates 4,000–4,750, then back to 2\texttimes{} & observation-originated, time-step-specific \\
5,000 & \texttt{updater.py} & Scheduled a test of Muon momentum 0.98, instead of 0.95, for updates 5,500–6,250 & observation-originated, time-step-specific \\
5,750 & \texttt{updater.py} & Cut the momentum test short, back to 0.95 from update 6,001, as the loss had risen & fix or revert, observation-originated, time-step-specific \\
6,500 & \texttt{updater.py} + \texttt{analysis.py} & Moved the two input embedders (1\texttimes{}1 convolutions) from AdamW to Muon & block-specific, observation-originated \\
7,000 & \texttt{updater.py} & Moved the input embedders back to AdamW, after no gain & fix or revert, block-specific, observation-originated \\
7,250 & \texttt{updater.py} & Scheduled a test: the AdamW groups and the timestep embedder at 1\texttimes{} for updates 7,500–8,250, the rest at 2\texttimes{} & block-specific, observation-originated, time-step-specific \\
8,500 & \texttt{updater.py} & Scheduled a second dip, to 0.3 of the rate for updates 8,576–8,650, after the loss had stalled & observation-originated, time-step-specific \\
8,750 & \texttt{updater.py} & Scheduled a third test, 2.75\texttimes{} for updates 9,000–9,750; fixed a bug that would have broken a rerun of its schedule & fix or revert, observation-originated, time-step-specific \\
11,250 & \texttt{updater.py} & Started the cooldown at update 11,250 instead of 18,765, after the loss stalled at 2\texttimes{} & observation-originated, time-step-specific \\
\end{xltabular}}

\exhead{Goldfish (Amharic)}

{\fontsize{9}{10}\selectfont
\renewcommand{\arraystretch}{0.94}
\begin{xltabular}{\textwidth}{@{}>{\raggedleft\arraybackslash}p{1.1cm}L{2.2cm}X L{3.0cm}@{}}
\toprule
Step & File & What changed & Tags \\
\midrule
\endhead
\bottomrule
\endfoot
323 & \texttt{updater.py} + \texttt{analysis.py} & Moved the 48 block matrices to Muon at 0.02 (Q, K and V orthogonalized apart) and kept AdamW at 2e-3 on the rest; ramp to the peak at update 1,600, then a linear decay to 0 & block-specific, time-step-specific \\
646 & \texttt{analysis.py} & Added a wake-up: an early decision when a window's mean loss rises 0.25 above the best recent one & analysis only \\
1,615 & \texttt{updater.py} & Halved both peak rates (Muon 0.02 to 0.01, AdamW 2e-3 to 1e-3) and doubled Muon's weight decay to 0.02 & block-specific, observation-originated \\
6,460 & \texttt{analysis.py} & Fixed an off-by-one in the same-batch comparison: an epoch is 6,462 updates, not 6,463 & analysis only, fix or revert \\
6,783 & \texttt{updater.py} & Raised Muon momentum from 0.95 to 0.975 & block-specific, observation-originated \\
7,106 & \texttt{updater.py} & Put Muon momentum back to 0.95, after the loss rose and the weights grew & fix or revert, block-specific, observation-originated \\
8,075 & \texttt{updater.py} & Lowered the Muon peak rate from 0.01 to 0.007 & block-specific, observation-originated \\
9,044 & \texttt{updater.py} & Cut the AdamW rate (embeddings, LayerNorms, biases) to 0.7\texttimes{} & block-specific, observation-originated \\
11,628 & \texttt{updater.py} & Raised that AdamW multiplier from 0.7 to 0.85, by the rule it had set at decision 28 & fix or revert, block-specific, observation-originated \\
12,597 & \texttt{analysis.py} & Added the loss over the last 323 updates minus the same updates one epoch earlier, as batch difficulty repeats each epoch & analysis only, observation-originated \\
\end{xltabular}}

\exhead{NanoGPT, \textonequarter{} batch}

{\fontsize{9}{10}\selectfont
\renewcommand{\arraystretch}{0.94}
\begin{xltabular}{\textwidth}{@{}>{\raggedleft\arraybackslash}p{1.1cm}L{2.2cm}X L{3.0cm}@{}}
\toprule
Step & File & What changed & Tags \\
\midrule
\endhead
\bottomrule
\endfoot
260 & \texttt{updater.py} + \texttt{analysis.py} & Rewrote the optimizer: Muon on the 72 block matrices on every rank, momentum restarted; rates, momenta and Muon weight decay moved part way to small-batch values &  \\
1,560 & \texttt{updater.py} & Gave the final norm gain, which scales the logits, its own rate, cut from 0.009 to 0.003 & block-specific, observation-originated \\
1,820 & \texttt{updater.py} + \texttt{analysis.py} & Put that gain's rate back to 0.009; added gradient-consistency measures from the optimizer state & fix or revert, block-specific, observation-originated \\
2,080 & \texttt{updater.py} & Lowered Muon's rate on the attention value matrices to 0.75\texttimes{} of the rest & block-specific, observation-originated \\
8,060 & \texttt{updater.py} & Made Muon's weight decay follow the learning-rate schedule from here on & niche practice, block-specific, observation-originated, time-step-specific \\
\end{xltabular}}

\end{document}